\PassOptionsToPackage{table}{xcolor}
\documentclass[sigconf]{acmart}

\usepackage{pifont}
\usepackage{multirow, multicol}
\usepackage{makecell}
\usepackage{algorithmic}
\usepackage[ruled,vlined]{algorithm2e}
\usepackage{bbm}
\usepackage{subcaption}
\usepackage{graphicx}
\definecolor{sgray}{HTML}{EDEDED}

\AtBeginDocument{
  }

\copyrightyear{2024}
\acmYear{2024}
\setcopyright{acmlicensed}\acmConference[MM '24]{Proceedings of the 32nd ACM International Conference on Multimedia}{October 28-November 1, 2024}{Melbourne, VIC, Australia}
\acmBooktitle{Proceedings of the 32nd ACM International Conference on Multimedia (MM '24), October 28-November 1, 2024, Melbourne, VIC, Australia}
\acmDOI{10.1145/3664647.3681206}
\acmISBN{979-8-4007-0686-8/24/10}

\begin{document}

\title[Cross-View Consistency Regularisation for Knowledge Distillation]{Cross-View Consistency Regularisation for \\ Knowledge Distillation}

\author{Weijia Zhang}
\affiliation{
  \institution{Shanghai Jiao Tong University}
  \city{Shanghai}
  \country{China}}
\email{weijia.zhang@sjtu.edu.cn}

\author{Dongnan Liu}
\affiliation{
  \institution{University of Sydney}
  \city{Sydney}
  \country{Australia}}
\email{dongnan.liu@sydney.edu.au}

\author{Weidong Cai}
\affiliation{
  \institution{University of Sydney}
  \city{Sydney}
  \country{Australia}}
\email{tom.cai@sydney.edu.au}

\author{Chao Ma}
\authornote{Corresponding author.}
\affiliation{
  \institution{Shanghai Jiao Tong University}
  \city{Shanghai}
  \country{China}}
\email{chaoma@sjtu.edu.cn}

\begin{abstract}
Knowledge distillation (KD) is an established paradigm for transferring privileged knowledge from a cumbersome model to a lightweight and efficient one. In recent years, logit-based KD methods are quickly catching up in performance with their feature-based counterparts. However, previous research has pointed out that logit-based methods are still fundamentally limited by two major issues in their training process, namely overconfident teacher and confirmation bias. Inspired by the success of cross-view learning in fields such as semi-supervised learning, in this work we introduce within-view and cross-view regularisations to standard logit-based distillation frameworks to combat the above cruxes. We also perform confidence-based soft label mining to improve the quality of distilling signals from the teacher, which further mitigates the confirmation bias problem. Despite its apparent simplicity, the proposed Consistency-Regularisation-based Logit Distillation (CRLD) significantly boosts student learning, setting new state-of-the-art results on the standard CIFAR-100, Tiny-ImageNet, and ImageNet datasets across a diversity of teacher and student architectures, whilst introducing no extra network parameters. Orthogonal to on-going logit-based distillation research, our method enjoys excellent generalisation properties and, without bells and whistles, boosts the performance of various existing approaches by considerable margins. Our code and models are available at \url{https://github.com/arcaninez/crld}.
\end{abstract}



\maketitle

\begin{figure}[th] \centering
  \includegraphics[width=\linewidth]{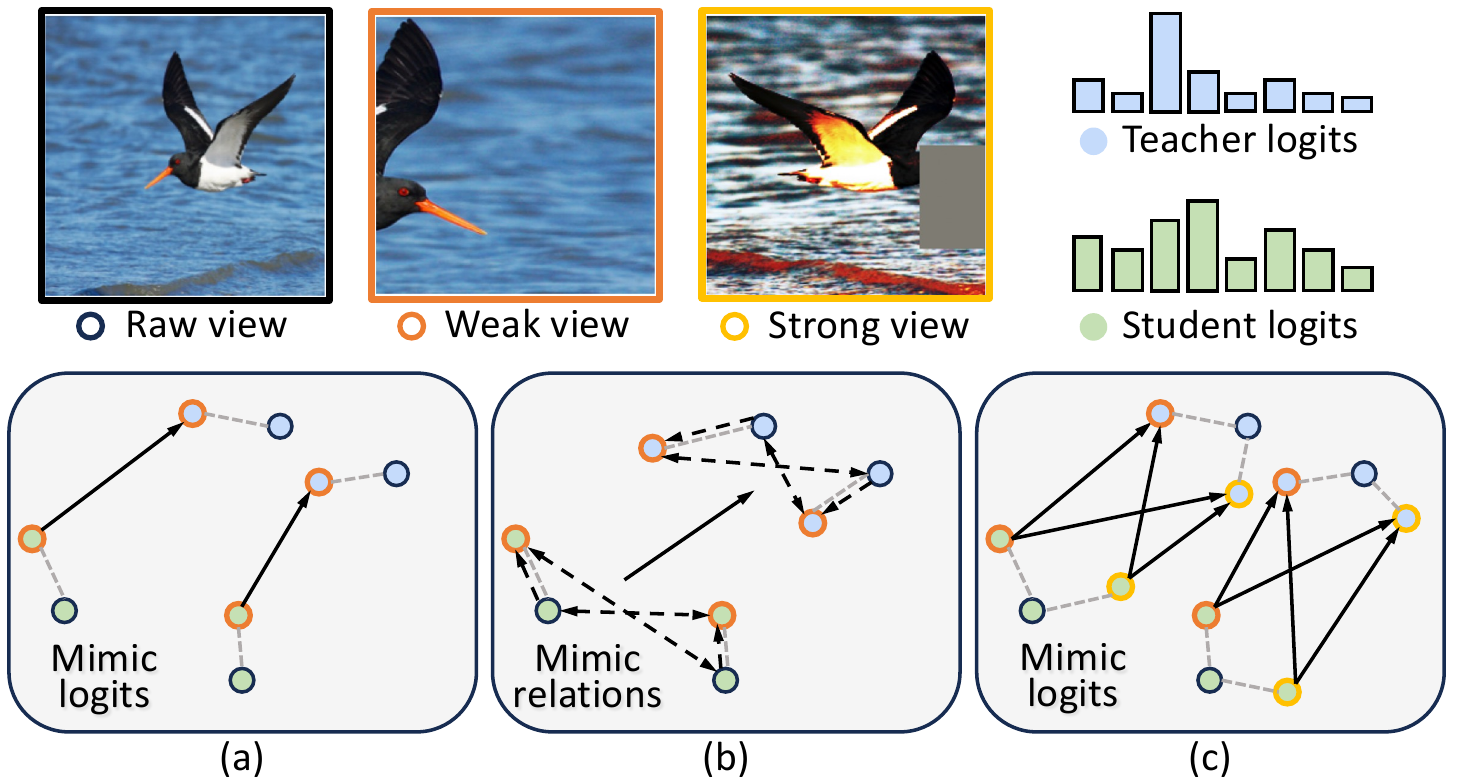}
  \caption{A schematic comparison of logit-based distillation methods from a cross-view learning perspective: (a) Methods mimicking logits in an unitary view \cite{kd, takd, ctkd, dkd, nkd, mlld, normkd}. (b) Methods optimising and mimicking contrastive relations \cite{sskd}. (c) The proposed CRLD which involves within-view and cross-view logit transfer.}
  \Description{A schematic comparison of logit-based distillation methods from a cross-view learning perspective.} \label{fig:concept} 
\end{figure}

\section{Introduction} 
Deep neural networks (DNNs) have achieved tremendous success across a plethora of computer vision, natural language processing, and multimedia tasks~\cite{vgg, resnet, vit}. Behind their widespread applications, high-performance DNNs are often associated with larger if not prohibitive model sizes and computational overheads, which render them hard to implement on resource-constrained devices and platforms. Towards computation-efficient, storage-friendly, and real-time deployment of DNNs, a viable solution is knowledge distillation (KD), which was first proposed by Hinton et al. \cite{kd} for model compression. KD works by transferring the advanced capability of a larger, cumbersome teacher model to a more lightweight and efficient student model. Since its proposal, KD has witnessed significant advancements in the past decade as a range of feature-based ~\cite{fitnets, ofd, crd, at, catkd, rkd, reviewkd, semckd, mgd, ickd, giftkd, spkd} and response-based (logit-based) \cite{kd, sskd, takd, dkd, mlld, nkd, normkd} KD algorithms are proposed for diverse tasks and applications \cite{kd, fgd, fgfi, cmkd, unidistill, structuredkd}. State-of-the-art KD methods have largely reduced the teacher-student performance gap. For instance, top-performing methods \cite{dkd, normkd, mlld} are capable of training students that are on par with or even surpass their corresponding teacher models on smaller datasets such as CIFAR-100 (see Table~\ref{tab:cifar100_homo}), and are not far behind on larger datasets \cite{imagenet} (Tables~\ref{tab:timagenet} and ~\ref{tab:imagenet}). 

In this paper, our goal is to further advance the capability of knowledge distillation by addressing two long-standing problems in existing KD methods. Previous research has reported that stronger teacher models and more accurate teacher predictions do not necessarily lead to better distilled students \cite{kd, eskd, takd, tfkd}. This counter-intuitive observation points to a prominent and fundamental problem in knowledge distillation --- \textbf{overconfident teacher}. In the pioneering work of KD \cite{kd}, Hinton et al. argued that valuable information is hidden in teacher's predictions of the non-target classes. These predictions, known as the ``dark information'', are however largely suppressed when the teacher make predictions with an overly-high confidence. Hence, regularisation of teacher predictions is essential to distilling knowledge with greater generalisation capabilities to the student \cite{nncalibration, label_smoothing, kd}. 

In their work \cite{kd}, Hinton et al. propose to mitigate the overconfidence problem by softening the predicted probabilities after Softmax using the temperature hyperparameter. This practice is inherited by many later works \cite{ctkd, dkd, normkd, nkd, mlld, lskd}. Some methods \cite{takd} produce smoothed teacher predictions by introducing auxiliary teacher networks with smaller capacity. More straightforward techniques have also been investigated, including label smoothing \cite{label_smoothing} and early stopping \cite{eskd}. These works also highlighted overfitting as another detrimental phenomenon closely related to overconfident teacher. These efforts motivate us to look at consistency regularisation via view transformation --- another viable solution to combat overconfidence and overfitting. Although widely explored in the semi-supervised learning (SSL) literature \cite{uda, fixmatch}, consistency regularisation and view transformation have received little attention in knowledge distillation research. According to \cite{unimatch}, strong augmentation amplifies the dark information that is insignificant in the weak view. As such, in this paper we reframe these techniques for KD by designing a novel set of within-view and cross-view consistency regularisation objectives and achieve state-of-the-art KD performance. 

On the other hand, teacher's predictions are not always correct. \textbf{Confirmation bias} \cite{confirmbias} arises when erroneous pseudo-labels predicted by the teacher is used to teach the student. 
In existing logit-based methods \cite{kd,ctkd,dkd,nkd,normkd}, the student is designated to faithfully learn whatever supervision the teacher has to provide. Such blind mimicking neglects a key fact that the teacher's predictions may be erroneous and misleading, thereby exacerbating the confirmation bias phenomenon.
It has been pointed out in recent research \cite{unimatch} that strong perturbation helps mitigate such confirmation bias, which also supports our introduction of a strongly-augmented view of the input image.
To further mitigate confirmation bias, we draw inspiration from state-of-the-art SSL frameworks whose success is partially attributed to their confidence-aware pseudo-labelling \cite{fixmatch,uda,unimatch}.  As such, we propose to selectively pick the more reliable predictions made by the teacher for the student to learn, which is proven beneficial in our experiments. 

The considerations and designs described above altogether lead to a novel \textit{C}onsistency-\textit{R}egularisation-based \textit{L}ogit \textit{D}istillation framework, dubbed ``CRLD''. By drawing inspiration and reaping the fruits from orthogonal research on semi-supervised learning (SSL), CRLD presents a simple yet highly effective and versatile solution to knowledge distillation. Besides reporting state-of-the-art results across different datasets and distillation pairs, CRLD also easily boosts advanced logit-based methods \cite{dkd, nkd, mlld, normkd} by considerable margins without introducing extra network parameters. A schematic comparison of CRLD against prior logit-based approaches from a cross-view learning perspective is depicted in Figure~\ref{fig:concept}.

In summary, the contributions of this paper include:
\begin{enumerate}
    \item We introduce extensive within-view and cross-view consistency regularisations to combat the overconfident teacher and over-fitting problems common in KD.  
    \item We design a reliable pseudo-label mining module to sidestep the negative impact of unreliable and erroneous supervisory signals from the teacher, thereby mitigating confirmation bias in KD.  
    \item We present the simple, versatile, and highly effective CRLD framework. CRLD achieves new state-of-the-art results on multiple benchmarks across diverse network architectures and readily boosts existing logit-based methods by considerable margins.
\end{enumerate}

\begin{figure*}[ht] \centering
  \includegraphics[height=5.2cm]{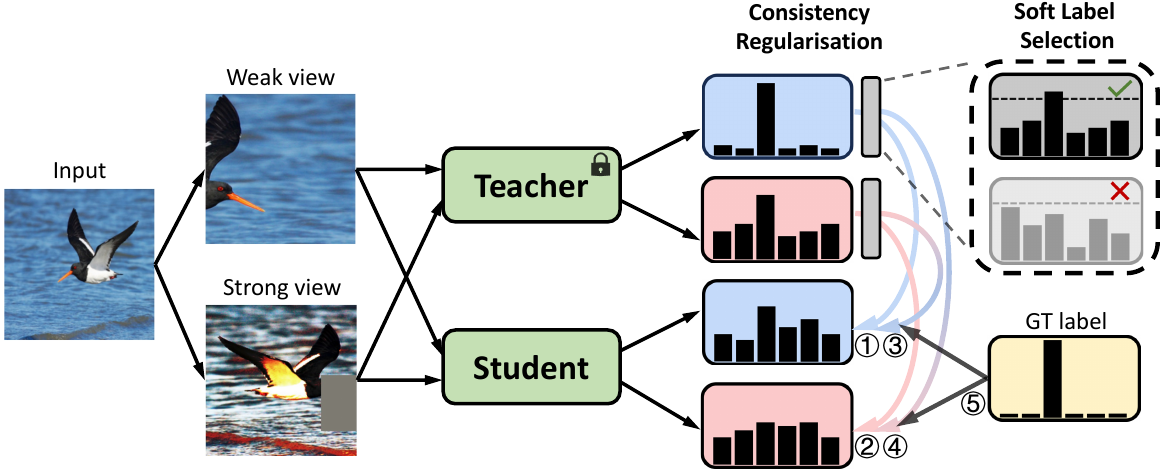}
  \caption{The CRLD framework. An input image is transformed into a weakly-transformed view and a strongly-transformed view. Both views are fed into the teacher and the student separately, yielding four predictions of the same instance. Amongst them, two types of consistency regularisation are enforced: within-view (\ding{172}\ding{173}) and cross-view (\ding{174}\ding{175}). Besides, student's predictions are supervised by ground-truths (\ding{176}) as per standard practice.}
  \label{fig:framework}
\end{figure*}

\section{Related Work}
\subsection{Knowledge Distillation}
Knowledge distribution (KD) is first proposed in \cite{kd} as a model compression technique. It transfers advanced knowledge from a larger, cumbersome ``teacher'' model to a smaller, lightweight ``student'' model. Following its nascent success in image classification \cite{kd, fitnets} and object detection \cite{fgfi, fgd, ld}, KD quickly has its effectiveness proven in more challenging downstream tasks \cite{ligastereo, cmkd, odm3d, unidistill, discofos, ekd, fretal, kddlgan, minillm}. Existing KD methods are primarily divided into feature-based and logit-based distillation according to where in the network knowledge transfer takes place --- the feature space or the logit space.

\noindent \textbf{Feature-based Distillation}. As its name suggests, feature-based distillation transfers knowledge in the intermediate feature space of the teacher and student models. A most straightforward way is to simply let the student mimic the features of the teacher, as is done in many early works \cite{fitnets, ofd, vid, crd}. 
Some methods also mine and transfer higher-order information from the teacher's feature maps to the student, including inter-channel \cite{ickd}, inter-layer \cite{giftkd}, inter-class \cite{dist}, intra-class \cite{dist}, and inter-sample \cite{pkt, cckd, rkd, spkd} correlations, as well as the teacher network's attention \cite{at, catkd}. 
Generative modelling has also been leveraged for feature-based distillation \cite{mgd}, where randomly masked student features are required to re-generate full teacher features.
In addition, cross-stage distillation paths \cite{reviewkd, semckd} and one-to-all pixel paths \cite{tat} are also proposed for improved feature-based distillation. 

\noindent \textbf{Logit-based Distillation. }
Logits are the prediction output by a neural network before its final Softmax layer. Distillation methods that perform knowledge transfer in the prediction space are referred to as logit-based or response-based distillation. Pioneering methods such as KD \cite{kd} and DML \cite{dml} directly transfer the teacher's predictions to the student by minimising the Kullback-Leibler (KL) divergence between their predictions. Akin to advances in feature-based distillation, intra-sample and inter-sample relations are also exploited for transfer within the logit space in several works \cite{sskd, mlld}.
Instead of treating all logits indiscriminately, DKD \cite{dkd} and NKD \cite{nkd} decompose all logits into target-class and non-target class logits and treat them separately, demonstrating stronger knowledge transfer performance.
More recent methods such as CTKD \cite{ctkd}, NormKD \cite{normkd}, LSKD~\cite{lskd}, and TTM~\cite{ttm} dynamically adjust the logit distribution, as opposed to fixed ones in previous works \cite{kd, dml, takd, dkd}, and have reported state-of-the-art performance. Another branch of methods \cite{takd, dgkd} introduce assistant networks between the teacher and the student to aid the imparting of logit-space knowledge to the latter. 

\subsection{Consistency Regularisation}
Consistency regularisation is at the core of the success of recent state-of-the-art semi-supervised learning algorithms \cite{temporalensembling, stochastic_transform, uda, meanteacher, mixmatch, remixmatch, fixmatch}. It involves enforcing invariant representations across different views of the same unlabelled input image to improve the generalisation of learnt representations on unseen data and distribution. The different views of an input image are generated by semantic-preserving transformations, from simple operations such as random crop, horizontal flip, and MixUp \cite{mixup} as weak transformations, to more sophisticated \cite{randaugment, augmix} or even adaptive \cite{autoaugment, remixmatch} augmentation strategies for producing strongly-augmented views. Given these artificially generated views, representation consistency can be enforced across two stochastical weak views as is done in \cite{mixmatch, uda}, or a pair of strong and weak views as in \cite{fixmatch, remixmatch}. To our best knowledge, beyond SSL, the idea of cross-view consistency regularisation has not been explored within the context of knowledge distillation.

\subsection{Data Augmentation for KD}
Data augmentation has been a pillar of deep learning's decade-long triumph. By transforming training samples into augmented versions whilst preserving their semantic connotation, data augmentation conveniently produces an abundant if not unlimited amount of extra training data to improve the generalisation of deep neural networks. In the context of knowledge distillation, data augmentation is yet to receive considerable attention, with only a handful of preliminary studies conducted \cite{dakd, funmatch, cdscore, ida}. Specifically, Das et al. \cite{cdscore} study the effect of data augmentation in training the teacher model. Wang et al. \cite{dakd} and IDA \cite{ida} design data augmentation strategies tailored to the KD task. SSKD \cite{sskd} and HSAKD \cite{hsakd} incorporate elements of contrastive learning. They apply simple rotation to establish a self-supervised pretext task for improved student learning. Our focus in this work is not the design of a data augmentation strategy itself but to leverage the idea of consistency regularisation to improve student's learning. In other words, data augmentation is simply our tool, which makes the formulation of consistency regularisation objectives possible, as is done in advanced SSL methods. 

\section{Methodology}
\subsection{Knowledge Distillation}
Knowledge distillation (KD) involves the student model learning from both ground-truth (GT) labels and distillation signals from a pre-trained teacher. For the image classification task, GT supervisions are widely enforced via a cross-entropy minimisation objective $\mathcal{L}_{CE}$; the distillation objective $\mathcal{L}_{KD}$ is enforced by minimising the distance between either the intermediate features or the final predictions of the teacher and the student. Thus, KD in its simplest form has $\mathcal{L} = \mathcal{L}_{CE}+\lambda_{KD} \mathcal{L}_{KD}$ as its objective, where $\lambda_{KD}$ is a balancing scalar.
In this paper, we investigate logit-based distillation, where $\mathcal{L}_{KD}$ minimises the discrepancy between predicted probabilities by the teacher and the student, and is commonly implemented as the Kullback-Leibler divergence (KLD) loss.

\begin{algorithm}[t] \small
\SetAlgoLined \SetKwInOut{Input}{Input}
\Input{A batch of training samples $\mathbf{x}$ \& their labels $\mathbf{y}$; weak augmentation $T_w(\cdot)$ \& strong augmentation $T_s(\cdot)$; teacher network $\mathcal{F}^T$ with parameters  $\theta^T$ \& student network $\mathcal{F}^S$ with parameters $\theta^S$} 
\While{model $ \mathcal{F}^S $ not converged} {
    \For{i=1 to step}{
        $\mathbf{p}^T_w = \mathcal{F}^T(T_w(\mathbf{x}); \theta^T) \quad\quad \mathbf{p}^T_s = \mathcal{F}^T(T_s(\mathbf{x}); \theta^T)$ \\ 
        $\mathbf{p}^S_w = \mathcal{F}^S(T_w(\mathbf{x}); \theta^S) \quad\quad \mathbf{p}^S_s = \mathcal{F}^S(T_s(\mathbf{x}); \theta^S)$ \\
        
        $\mathbf{M}_w = \text{SLS}(\mathbf{p}^T_w, \tau_w) \quad\quad\quad \mathbf{M}_s = \text{SLS}(\mathbf{p}^T_s, \tau_s)$ \\
        $\mathcal{L}_{CE} = \text{CE}(\mathbf{p}^S_w, \mathbf{y}) + \text{CE}(\mathbf{p}^S_s, \mathbf{y}) $ \\ 
        
        $\mathcal{L}^{WV}_{KD} = \text{KLD}( \mathbf{p}^S_w, \mathbf{p}^T_w) \mathbf{M}_w + \text{KLD}(\mathbf{p}^S_s, \mathbf{p}^T_s) \mathbf{M}_s $\\ 
        
        $\mathcal{L}^{CV}_{KD} = \text{KLD}( \mathbf{p}^S_s, \mathbf{p}^T_w) \mathbf{M}_w + \text{KLD}(\mathbf{p}^S_w, \mathbf{p}^T_s) \mathbf{M}_s $ \\ 
        
        $\mathcal{L}_{KD} = \mathcal{L}^{WV}_{KD} + \mathcal{L}^{CV}_{KD} $ \\ 
        
        $\mathcal{L}_{Overall} = \mathcal{L}_{CE} + 
        \lambda_{KD} \mathcal{L}_{KD} $ \\ 
        Update $\theta^S$ acc. to $\mathcal{L}_{Overall}$
    }
}
\SetKwInOut{Output}{Output}
\Output{Well-trained model $\mathcal{F}^S$ with parameters $\theta^S$}
\caption{The CRLD algorithm} \label{alg:crld} 
\end{algorithm}

\begin{table*}[th] \centering \small \tabcolsep=0.22cm
\caption{Top-1 accuracy (\%) on CIFAR-100 with homogeneous-architecture teacher-student pairs.}
\renewcommand{\arraystretch}{0.90}
\begin{tabular}{cc|cccccc|c} \toprule
\multirow{4}{*}{\begin{tabular}[c]{@{}c@{}} Method \\ \end{tabular}} & \multirow{2}{*}{Teacher}  & ResNet56 & ResNet110 & ResNet32$\times$4 & WRN-40-2 & WRN-40-2 & VGG13 &
\multirow{4}{*}{\begin{tabular}[c]{@{}c@{}} Avg. \\ \end{tabular}} \\
&  & 72.34 & 74.31 & 79.42 & 75.61 & 75.61 & 74.64 \\
& \multirow{2}{*}{Student} & ResNet20 & ResNet32 & ResNet8$\times$4 & WRN-16-2 & WRN-40-1 & VGG8 \\
& \space  & 69.06 & 71.14 & 72.50 & 73.26 & 71.98 & 70.36 & \\
\Xhline{3\arrayrulewidth} 
\multirow{12}{*}{Feature KD}
& RKD \cite{rkd} & 69.61 & 71.82 & 71.90 & 73.35 & 72.22 & 71.48 & 71.73 \\
& FitNets \cite{fitnets} & 69.21 & 71.06 & 73.50 & 73.58 & 72.24 & 71.02 & 71.77 \\
& AT \cite{at} & 70.55 & 72.31 & 73.44 & 74.08 & 72.77 & 71.43 & 72.43 \\
& OFD \cite{ofd} & 70.98 & 73.23 & 74.95 & 75.24 & 74.33 & 73.95 & 73.78\\
& CRD \cite{crd} & 71.16 & 73.48 & 75.51 & 75.48 & 74.14 & 73.94 & 73.95 \\
& SRRL \cite{srrl} & 71.13 & 73.48 & 75.33 & 75.59 & 74.18 & 73.44 & 73.86 \\
& ICKD \cite{ickd} & 71.76 & 73.89 & 75.25 & 75.64 & 74.33 & 73.42 & 74.05 \\
& PEFD \cite{pefd} & 70.07 & 73.26 & 76.08 & 76.02 & 74.92 & 74.35 & 74.12 \\
& CAT-KD \cite{catkd} & 71.05 & 73.62 & 76.91 & 75.60 & 74.82 & 74.65 & 74.44 \\
& TaT \cite{tat} & 71.59 & 74.05 & 75.89 & 76.06 & 74.97 & 74.39 & 74.49\\
& ReviewKD \cite{reviewkd} & 71.89 & 73.89 & 75.63 & 76.12 & 75.09 & 74.84 & 74.58 \\
& SimKD \cite{simkd} & 71.05 & 73.92 & 78.08 & 75.53 & 74.53 & 74.89 & 74.67 \\
\hline 
\multirow{13}{*}{Logit KD}       
& KD \cite{kd} & 70.66 & 73.08 & 73.33 & 74.92 & 73.54 & 72.98 & 73.09 \\
& TAKD \cite{takd} & 70.83 & 73.37 & 73.81 & 75.12 & 73.78 & 73.23 & 73.36 \\
& CTKD \cite{ctkd} & 71.19 & 73.52 & 73.79 & 75.45 & 73.93 & 73.52 & 73.57 \\
& NKD \cite{nkd} & 70.40 & 72.77 & 76.35 & 75.24 & 74.07 & 74.86 & 73.95 \\
& TTM \cite{ttm} & 71.83 & 73.97 & 76.17 & 76.23 & 74.32 & 74.33 & 74.48 \\
& LSKD \cite{lskd} & 71.43 & 74.17 & 76.62 & 76.11 & 74.37 & 74.36 & 74.51 \\
& NormKD \cite{normkd} & 71.40 & 73.91 & 76.57 & 76.40 & 74.84 & 74.45 & 74.60 \\
& DKD \cite{dkd} & 71.97 & 74.11 & 76.32 & 76.24 & 74.81 & 74.68 & 74.69\\
& \cellcolor{sgray} \textbf{CRLD} & \cellcolor{sgray} \textbf{72.10} & \cellcolor{sgray} \textbf{74.42} & \cellcolor{sgray} \textbf{77.60} & \cellcolor{sgray} \textbf{76.45} & \cellcolor{sgray} \textbf{75.58} & \cellcolor{sgray} \textbf{75.27} & \cellcolor{sgray} \textbf{75.24} \\ 
& \cellcolor{sgray} \textbf{CRLD-NormKD} & \cellcolor{sgray} \textbf{72.08} & \cellcolor{sgray} \textbf{74.59} & \cellcolor{sgray} \textbf{78.22} & \cellcolor{sgray} \textbf{76.49} & \cellcolor{sgray} \textbf{75.71} & \cellcolor{sgray} \textbf{75.48} & \cellcolor{sgray} \textbf{75.43} \\ \cline{2-9}
& MLLD $\dagger$ \cite{mlld} & 72.19 & 74.11 & 77.08 & 76.63 & 75.35 & 75.18 & 75.09  \\
& \cellcolor{sgray} \textbf{CRLD} $\dagger$ & \cellcolor{sgray} \textbf{72.42} & \cellcolor{sgray} \textbf{74.87} & \cellcolor{sgray} \textbf{78.28} & \cellcolor{sgray} \textbf{76.94} & \cellcolor{sgray} \textbf{76.02} & \cellcolor{sgray} \textbf{75.45} & \cellcolor{sgray} \textbf{75.66} \\
& \cellcolor{sgray} \textbf{CRLD-NormKD} $\dagger$ & \cellcolor{sgray} \textbf{72.57} & \cellcolor{sgray} \textbf{75.08} & \cellcolor{sgray} \textbf{78.53} & \cellcolor{sgray} \textbf{76.91} & \cellcolor{sgray} \textbf{76.20} & \cellcolor{sgray} \textbf{75.59} & \cellcolor{sgray} \textbf{75.81} \\
\bottomrule \label{tab:cifar100_homo} \vspace{-3pt}
\end{tabular}
\end{table*}

\begin{table*}[bt] \centering \small \tabcolsep=0.12cm
\caption{Top-1 accuracy (\%) on CIFAR-100 with heterogeneous-architecture teacher-student pairs.} 
\renewcommand{\arraystretch}{0.90}
\begin{tabular}{cc|cccccc|c}
\toprule \tabcolsep=0.02cm
\multirow{4}{*}{\begin{tabular}[c]{@{}c@{}} Method \\ \end{tabular}} & \multirow{2}{*}{Teacher} & ResNet32$\times$4 & ResNet32$\times$4 & WRN-40-2 & WRN-40-2 & VGG13 & ResNet50 &
\multirow{4}{*}{\begin{tabular}[c]{@{}c@{}} Avg. \\ \end{tabular}} \\
& & 79.42 & 79.42 & 75.61 & 75.61 & 74.64 & 79.34 \\
& \multirow{2}{*}{Student} & ShuffleNetV2 & WRN-16-2 & ResNet8$\times$4 & MobileNetV2 & MobileNetV2 & MobileNetV2\\
& & 71.82 & 73.26 & 72.50 & 64.60 & 64.60 & 64.60 & \\
\Xhline{3\arrayrulewidth} 
\multirow{8}{*}{Feature KD}
& AT \cite{at} & 72.73 & 73.91 & 74.11 & 60.78 & 59.40 & 58.58 & 66.59 \\
& RKD \cite{rkd} & 73.21 & 74.86 & 75.26 & 69.27 & 64.52 & 64.43 & 70.26 \\
& FitNets \cite{fitnets} & 73.54 & 74.70 & 77.69 & 68.64 & 64.16 & 63.16 & 70.32 \\
& CRD \cite{crd} & 75.65 & 75.65 & 75.24 & 70.28 & 69.63 & 69.11 & 72.59 \\
& OFD \cite{ofd} & 76.82 & 76.17 & 74.36 & 69.92 & 69.48 & 69.04 & 72.63 \\
& ReviewKD \cite{reviewkd} & 77.78 & 76.11 & 74.34 & 71.28 & 70.37 & 69.89 & 73.30 \\
& SimKD \cite{simkd} & 78.39 & 77.17 & 75.29 & 70.10 & 69.44 & 69.97 & 73.39 \\
& CAT-KD \cite{catkd} & 78.41 & 76.97 & 75.38 & 70.24 & 69.13 & 71.36 & 73.58 \\ \hline 
\multirow{10}{*}{Logit KD}      
& KD \cite{kd} & 74.45 & 74.90 & 73.97 & 68.36 & 67.37 & 67.35 & 71.07 \\
& CTKD \cite{ctkd} & 75.37 & 74.57 & 74.61 & 68.34 & 68.50 & 68.67 & 71.68 \\
& LSKD \cite{lskd} & 75.56 & 75.26 & 77.11 & 69.23 & 68.61 & 69.02 & 72.47 \\
& NormKD \cite{normkd} & 76.01 & 75.17 & 76.80 & 69.14 & 69.53 & 69.57 & 72.70 \\
& DKD \cite{dkd} & 77.07 & 75.70 & 75.56 & 69.28 & 69.71 & 70.35 & 72.95 \\
&  \cellcolor{sgray} \textbf{CRLD} &  \cellcolor{sgray} \textbf{78.27} &  \cellcolor{sgray} \textbf{76.92} &  \cellcolor{sgray} \textbf{77.28} & \cellcolor{sgray} \textbf{70.37} &  \cellcolor{sgray} \textbf{70.39} &  \cellcolor{sgray} \textbf{71.36} &  \cellcolor{sgray} \textbf{74.10} \\
&  \cellcolor{sgray} \textbf{CRLD-NormKD} &  \cellcolor{sgray} \textbf{78.35} &  \cellcolor{sgray} \textbf{77.28} &  \cellcolor{sgray} \textbf{77.51} & \cellcolor{sgray} \textbf{70.55} &  \cellcolor{sgray} \textbf{70.34} &  \cellcolor{sgray} \textbf{71.47} &  \cellcolor{sgray} \textbf{74.25} \\
\cline{2-9}
& MLLD $\dagger$ \cite{mlld} & 78.44 & 76.52 & 77.33 & 70.78 & 70.57 & 71.04 & 74.11 \\
&  \cellcolor{sgray} \textbf{CRLD $\dagger$} &  \cellcolor{sgray} \textbf{78.50} & \cellcolor{sgray} \textbf{77.04} &  \cellcolor{sgray} \textbf{77.75} &  \cellcolor{sgray} \textbf{71.26} &  \cellcolor{sgray} \textbf{70.70} & \cellcolor{sgray} \textbf{71.43} &  \cellcolor{sgray} \textbf{74.45} \\
&  \cellcolor{sgray} \textbf{CRLD-NormKD $\dagger$} &  \cellcolor{sgray} \textbf{78.52} & \cellcolor{sgray} \textbf{77.39} &  \cellcolor{sgray} \textbf{77.98} &  \cellcolor{sgray} \textbf{71.36} &  \cellcolor{sgray} \textbf{70.81} & \cellcolor{sgray} \textbf{71.49} &  \cellcolor{sgray} \textbf{74.59} \\
\bottomrule \label{tab:cifar100_het} \vspace{-10pt}
\end{tabular}
\end{table*}

\begin{table}[t] \centering \small
\caption{Top-1 and Top-5 accuracy (\%) on Tiny-ImageNet.} 
\renewcommand{\arraystretch}{0.90}
\begin{tabular}{cc|c} \toprule
\multirow{4}{*}{\begin{tabular}[c]{@{}c@{}} Method \\ \end{tabular}} & \multirow{2}{*}{Teacher} & ResNet32$\times$4 \\
& &  64.30\slash85.07 \\
& \multirow{2}{*}{Student} & ResNet8$\times$4 \\
& & 55.25\slash79.62 \\
\Xhline{3\arrayrulewidth} 
\multirow{1}{*}{Feature KD}
 & FCFD \cite{fcfd} & 60.15\slash82.80  \\
 \hline 
\multirow{9}{*}{Logit KD}    
& KD \cite{kd} & 56.00\slash79.64 \\
& DKD \cite{dkd} & 57.79\slash81.57 \\
& NKD \cite{nkd} & 58.63\slash82.12 \\
& NormKD \cite{normkd} & 62.05\slash83.98 \\
& \cellcolor{sgray} \textbf{CRLD} & \cellcolor{sgray} \textbf{63.39}\slash\textbf{84.20} \\
& \cellcolor{sgray} \textbf{CRLD-NormKD} & \cellcolor{sgray} \textbf{63.77}\slash\textbf{84.57} \\ \cline{2-3}
& MLLD $\dagger$ \cite{mlld} & 61.91\slash83.77 \\
& \cellcolor{sgray} \textbf{CRLD} $\dagger$ & \cellcolor{sgray} \textbf{63.65}\slash\textbf{84.74}  \\
& \cellcolor{sgray} \textbf{CRLD-NormKD} $\dagger$ & \cellcolor{sgray} \textbf{63.84}\slash\textbf{85.52}  \\
\bottomrule \label{tab:timagenet} \vspace{-5pt}
\end{tabular}
\end{table}

\subsection{Logit-Space Consistency Regularisation} \label{sec:consistency}
Consistency regularisation has been widely employed in SSL research \cite{stochastic_transform, meanteacher, fixmatch, mixmatch, remixmatch}. It involves creating different views of the same unlabelled image, which are separately fed into a neural network to obtain a pair of network predictions. 
Consistency regularisation is enforced between the pair of predictions given the prior knowledge that both views fundamentally represent the same high-level information such as the object category.

In CRLD, we employ one weak and one strong view to set the stage for our set of within-view and cross-view consistency criteria. Specifically, we adopt RandAugment \cite{randaugment} with random magnitude alongside random crop, random horizontal flip, and Cutout \cite{cutout} as our strong data augmentation policy. A full list of RandAugment's transformation operations is provided in the Supplementary Materials. For the weak augmentation, we simply apply random crop and random horizontal flip, which is the standard data augmentation in previous logit-based KD methods \cite{kd, dkd, normkd, nkd, lskd}. We denote our weak and strong view transformation functions by $T_w(\cdot)$ and $T_s(\cdot)$, respectively.

Concretely, given a batch of $B$ training samples $\mathbf{x} = \{ x_b : b \in (1,...,B) \}$, we separately apply $T_w(\cdot)$ and $T_s(\cdot)$ to each sample to obtain a weakly-augmented and a strongly-augmented view of $x_b$, denoted as $x^w_b$ and $x^s_b$, respectively. Next, we feed both views of the input image individually to the teacher and the student, obtaining four network predictions, namely $\mathbf{p}_w^T$, $\mathbf{p}_s^T$, $\mathbf{p}_w^S$, and $\mathbf{p}_s^S$, where we drop subscript $b$ for brevity.

We define within-view consistency regularisation as the consistency criterion between teacher's and student's predictions of the same weak or strong view. The within-view consistency objective is therefore computed as:
\begin{equation}
    \mathcal{L}^{WV}_{KD} = \text{KLD}( \mathbf{p}^S_w, \mathbf{p}^T_w) + \text{KLD}(\mathbf{p}^S_s, \mathbf{p}^T_s)
\end{equation}

Next, we design a novel cross-view consistency regularisation. It demands the teacher and student to receive differently augmented views of an image and yet produce logit predictions as similar as possible. Formally, this cross-view objective is given by: 
\begin{equation}
    \mathcal{L}^{CV}_{KD} = \text{KLD}( \mathbf{p}^S_w, \mathbf{p}^T_s) + \text{KLD}(\mathbf{p}^S_s, \mathbf{p}^T_w)
    \label{eqn:cv_loss}
\end{equation}

The overall KD objective is a weighted sum of the within-view and cross-view consistency losses: $\mathcal{L}_{KD} = \lambda^{WV}_{KD} \mathcal{L}^{WV}_{KD} + \lambda^{CV}_{KD} \mathcal{L}^{CV}_{KD}$. A schematic diagram of the pipeline is provided in Figure~\ref{fig:framework} .  

Furthermore, a previous work \cite{funmatch} reported that teacher and student shall receive an identical view of the same input using the same image transformation for maximal knowledge distillation performance. With our specific design, however, we will demonstrate that the teacher and student receiving different views of an input using different view transformations leads to optimal performance. In Section~\ref{sec:ablation}, we conduct extensive ablation experiments to examine whether and when the proposed cross-view learning really works. As will be shown, cross-view consistency regularisation using different views of the same input image is key to the strong performance of the proposed CRLD framework.

\subsection{Confidence-based Soft Label Mining}
Confirmation bias harms distillation when the student learns from erroneous soft labels provided by the teacher. By introducing the more challenging strongly-augmented views, we are also increasing the likelihood that the well-trained teacher produces misleading predictions that undermine student learning. We experimentally observe that strongly-augmented samples generated by our strong view transformation policy can sometimes be almost unintelligible (refer to Supplementary Materials for examples), with false predictions made by the teacher. 
Therefore, we are motivated to refrain unreliable teacher predictions from forming the consistency regularisation pairs. To this end, we propose a simple thresholding mechanism by considering the highest class probability in teacher's per-instance prediction as an indicator of teacher's uncertainty about this prediction. Teacher predictions whose highest class probability is below a given threshold are discarded. 

In practice, we apply two thresholds $\tau_w$ and $\tau_s$ for teacher's predictions of the weak and strong views, respectively. Different from the common practice in SSL \cite{uda, fixmatch}, we do not convert the preserved predictions into hard, one-hot pseudo-labels. This is due to the nature of the KD task, whose success hinges on the dark knowledge carried within the non-target class predictions \cite{kd,dkd,nkd}.
Instead, we keep teacher's soft predictions as they are as supervision. As an example, $\text{KLD}( \mathbf{p}^S_w, \mathbf{p}^T_s)$ in Equation \ref{eqn:cv_loss} becomes $ \mathbbm{1} (\max \mathbf{p}^T_s > \tau_s) \text{KLD}(\mathbf{p}^S_w, \mathbf{p}^T_s)$,
where $\mathbbm{1}(\cdot)$ is the indicator function. Other objectives are defined like-wise and are omitted for brevity.

\begin{table}[t] \centering \small \tabcolsep=0.12cm
\caption{Top-1 and Top-5 accuracy (\%) on ImageNet.} 
\renewcommand{\arraystretch}{0.90}
\begin{tabular}{cc|cc} \toprule \tabcolsep=0.02cm
\multirow{4}{*}{\begin{tabular}[c]{@{}c@{}} Method \\ \end{tabular}} & \multirow{2}{*}{Teacher} & ResNet34 & ResNet50 \\
&  & 73.31\slash91.42 & 76.16\slash92.86 \\
& \multirow{2}{*}{Student} & ResNet18 & MobileNetV1 \\
&  & 69.75\slash89.07 & 68.87\slash88.76 \\
\Xhline{3\arrayrulewidth} 
\multirow{8}{*}{Feature KD}
 & AT \cite{at} & 70.69\slash90.01 & 69.56\slash89.33 \\
 & OFD \cite{ofd} & 70.81\slash89.98 & 71.25\slash90.34 \\
 & CRD \cite{crd} & 71.17\slash90.13 & 71.37\slash90.41 \\
 & RKD \cite{rkd} & 71.34\slash90.37 & 71.32\slash90.62 \\
 & CAT-KD \cite{catkd} & 71.26\slash90.45 & 72.24\slash91.13 \\
 & SimKD \cite{simkd} & 71.59\slash90.48 & 72.25\slash90.86 \\
 & ReviewKD \cite{reviewkd} & 71.61\slash90.51 & 72.56\slash91.00 \\
 & SRRL \cite{srrl} & 71.73\slash90.60 & 72.49\slash90.92 \\
 \hline 
\multirow{11}{*}{Logit KD}      
& KD \cite{kd} & 70.66\slash89.88 & 68.58\slash88.98 \\
& TAKD \cite{takd} & 70.78\slash90.16 & 70.82\slash90.01 \\
& CTKD \cite{ctkd} & 71.51\slash- & 90.47\slash- \\
& NormKD \cite{normkd} & 71.56\slash90.47 & 72.12\slash90.86  \\
& DKD \cite{dkd} & 71.70\slash90.41 & 72.05\slash91.05 \\
& NKD \cite{nkd} & 71.96\slash- & 72.58\slash-  \\
& MLLD \cite{mlld} & 71.90\slash90.55 & 73.01\slash91.42  \\
& TTM \cite{ttm} & 72.19\slash- & 73.09\slash-  \\
& LSKD \cite{lskd} & 72.08\slash90.74 & 73.22\slash91.59 \\
& \cellcolor{sgray} \textbf{CRLD} & \cellcolor{sgray} \textbf{72.37}\slash\textbf{90.76} & \cellcolor{sgray} \textbf{73.53}\slash 91.43  \\
& \cellcolor{sgray} \textbf{CRLD-NormKD} & \cellcolor{sgray} \textbf{72.39}\slash\textbf{90.87} & \cellcolor{sgray} \textbf{73.74}\slash \textbf{91.61} \\
\bottomrule \label{tab:imagenet}
\end{tabular} \vspace{-8pt}
\end{table}

\subsection{Training Objective}
The overall training objective for CRLD is a weighted combination of previously described loss terms, namely a ground-truth supervision loss $\mathcal{L}_{CE}$ and a teacher supervision KD loss $\mathcal{L}_{KD}$.
\begin{equation}
    \mathcal{L} = \mathcal{L}_{CE} + \lambda_{KD} \mathcal{L}_{KD}  = \mathcal{L}_{CE} + \lambda^{WV}_{KD} \mathcal{L}^{WV}_{KD} + \lambda^{CV}_{KD} \mathcal{L}^{CV}_{KD}
\label{eqn:loss_crld}
\end{equation}
where $\lambda_{KD}$ is a balancing weight. $\mathcal{L}_{CE}$ is computed between student's predictions of both the weakly- and strong-augmented inputs and the GT label using the cross-entropy loss.

The pseudo-code for the training of CRLD is provided in Algorithm~\ref{alg:crld}, where $\text{SLS}(\cdot)$ denotes the confidence-based soft label mining operation with $\tau_{w}$ or $\tau_{s}$ as its parameter. $\text{SLS}(\cdot)$ produces binary mask $\mathbf{M}$ which indicates the selected instance-wise predictions.

\section{Experiments}
\subsection{Datasets} 
\noindent \textbf{CIFAR-100}~\cite{cifar100}: a classic image classification benchmark with 50{\small,}000 training and 10{\small,}000 validation (or test) RGB images of 100 classes.

\noindent \textbf{Tiny-ImageNet}: a subset of ImageNet \cite{imagenet} which consists of 100{\small,}000 training and 50{\small,}000 validation RGB images over 200 classes, with image resolution downsized from ImageNet's $256\times256$ to $64\times64$.

\noindent \textbf{ImageNet}~\cite{imagenet}: a widely used large-scale image classification dataset, comprising 1.28 million training and 50{\small,}000 validation RGB images annotated in 100 classes. 

\subsection{Implementation Details}
We evaluate our method across various teacher-student pairs of common DNN architecture families: ResNet~\cite{resnet}, WRN~\cite{wrn}, VGG~\cite{vgg}, MobileNet~\cite{mobilenet,mobilenetv2}, and ShuffleNet~\cite{shufflenet}. In all experiments, we strictly adhere to standardised training configurations of previous knowledge distillation methods \cite{kd, at, fitnets, ofd, spkd, cckd, pkt, semckd, crd, rkd, reviewkd, pefd, srrl, hsakd, normkd, nkd, ctkd, catkd}. All reported experimental results are averaged over 3 independent runs. 

Specifically, for CIFAR-100 and Tiny-ImageNet, we train our method for a total of 240 epochs, with an initial learning rate of 0.025 for MobileNet \cite{mobilenetv2} and ShuffleNet \cite{shufflenet} students and 0.05 for others. The learning rate decays by a factor of 10 after the 150th, 180th, and 210th epochs; the SGD optimiser is used, with a momentum of 0.9, a weight decay of $5\times10^{-4}$, and a batch size of 64. For ImageNet, we conduct 100-epoch training with a batch size of 512 and an initial learning rate of 0.2 that decays by a factor of 10 at the 30th, 60th, and 90th epochs. Other parameters, unless otherwise stated, follow CIFAR-100 and Tiny-ImageNet experiments. Our method is implemented in the \textit{mdistiller} codebase.

\begin{table}[t]
\caption{Ablation experiments on different consistency regularisation designs using CIFAR-100.}
\center \small \tabcolsep=0.15cm \captionsetup{skip=6pt}
\renewcommand{\arraystretch}{0.85}
  \begin{tabular}{c|cccc|c} \toprule
   \multirow{2}{*}{\begin{tabular}[c]{@{}c@{}} Expt. \\ \end{tabular}} & \multirow{2}{*}{$\mathbf{p}^S_w$ - $\mathbf{p}^T_w$} & \multirow{2}{*}{$\mathbf{p}^S_s$ - $\mathbf{p}^T_s$} & \multirow{2}{*}{$\mathbf{p}^S_w$ - $\mathbf{p}^T_s$} & \multirow{2}{*}{$\mathbf{p}^S_s$ - $\mathbf{p}^T_w$} & ResNet32$\times$4 \\
 & &  &  & & ResNet8$\times$4 \\ \Xhline{3\arrayrulewidth} 
A & \ding{52} & &  & & 76.26 \\
B & & \ding{52} & & & 76.75 \\
C & & & \ding{52} & & 74.10 \\
D & & & & \ding{52} & 75.38 \\ \hline
E & \ding{52} & & \ding{52} & & 76.60 \\
F & \ding{52} & & & \ding{52} & 77.36 \\ \hline
G & \ding{52} & \ding{52} & & & 77.39 \\
H & \ding{52} & \ding{52} & \ding{52} & & 77.71 \\
I & \ding{52} & \ding{52} & & \ding{52} & 78.11 \\
\rowcolor[HTML]{EDEDED}
J & \ding{52} & \ding{52} & \ding{52} & \ding{52} & \textbf{78.18} \\ \bottomrule
\end{tabular} \label{tab:abl}
\end{table}

\begin{figure}[t] \centering
    \includegraphics[width=0.98\linewidth]{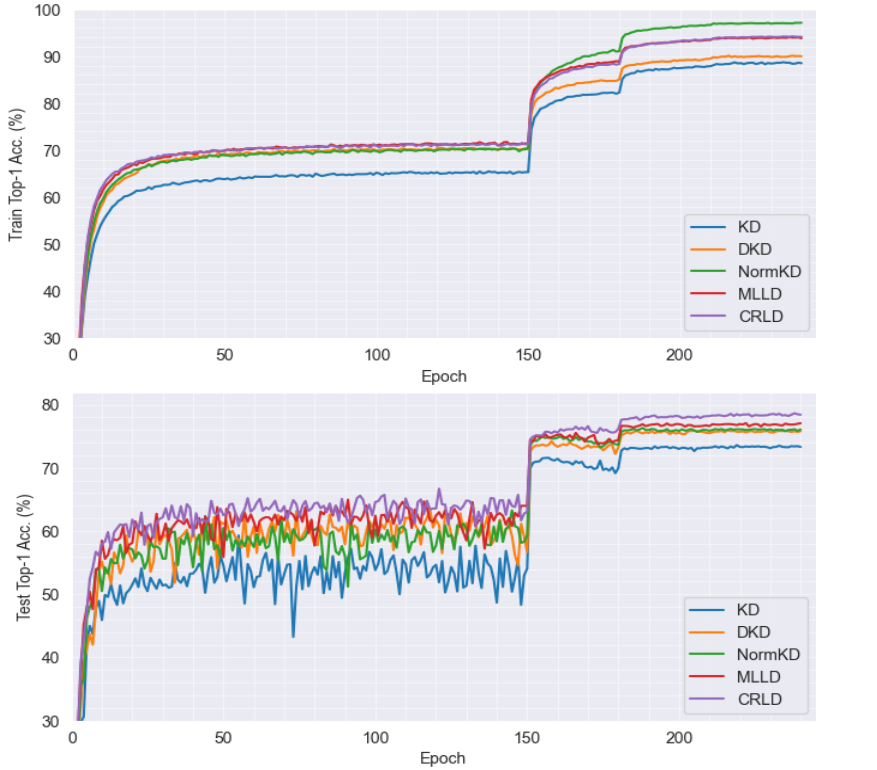} \vspace{-8pt} 
    \caption{Evolution of training (top) and test (bottom) set Top-1 accuracy (\%) on CIFAR-100.} \label{fig:acc_curve}  \vspace{-5pt} 
\end{figure}

\begin{table*}[ht] \centering \small
\caption{Generalisation to existing logit-based methods on CIFAR-100.}
\renewcommand{\arraystretch}{0.90}
\begin{tabular}{c|cccccc|c} \toprule
\multirow{2}{*}{Teacher}  & ResNet56 & ResNet110 & ResNet32$\times$4 & WRN-40-2 & WRN-40-2 & VGG13 &
\multirow{4}{*}{\begin{tabular}[c]{@{}c@{}} Avg. \\ \end{tabular}} \\
 & 72.34 & 74.31 & 79.42 & 75.61 & 75.61 & 74.64 \\
 \multirow{2}{*}{Student} & ResNet20 & ResNet32 & ResNet8$\times$4 & WRN-16-2 & WRN-40-1 & VGG8 \\
\space  & 69.06 & 71.14 & 72.50 & 73.26 & 71.98 & 70.36 & \\
\Xhline{3\arrayrulewidth} 
 KD \cite{kd} & 70.69 & 73.57 & 73.53 & 75.22 & 73.74 & 73.43 & 73.36 \\
 \rowcolor[HTML]{EDEDED}
\textbf{+CRLD} & \textbf{72.10} & \textbf{74.42} & \textbf{77.60} & \textbf{76.45} & \textbf{75.58} & \textbf{75.27} & \textbf{75.24} \\ \cline{1-8}
 NKD \cite{nkd} & 70.40 & 72.77 & 76.21 & 75.24 & 74.07 & 74.40 & 73.85 \\
 \rowcolor[HTML]{EDEDED}
 \textbf{+CRLD} & \textbf{71.95} & \textbf{74.40} & \textbf{78.16} & \textbf{76.60} & \textbf{74.87} & \textbf{75.16} & \textbf{75.19} \\ \cline{1-8}
 MLLD \cite{mlld} & 71.24 & 73.96 & 74.64 & 75.57 & 73.97 & 73.80 & 73.86 \\
 \rowcolor[HTML]{EDEDED}
 \textbf{+CRLD} & \textbf{72.07} & \textbf{74.64} & \textbf{77.00} & \textbf{76.75} & \textbf{75.46} & \textbf{74.87} & \textbf{75.13} \\ \cline{1-8}
 DKD \cite{dkd} & \textbf{71.49} & \textbf{73.95} & 75.96 & 75.67 & 74.47 & 74.67 & 74.37 \\
 \rowcolor[HTML]{EDEDED}
 \textbf{+CRLD} & 70.70 & 73.45 & \textbf{77.90} & \textbf{76.27} & \textbf{75.16} & \textbf{75.57} & \textbf{74.84} \\
 \cline{1-8}
 NormKD \cite{normkd} & 71.43 & 73.95 & 76.26 & 76.01 & 74.55 & 74.45 & 74.44 \\
 \rowcolor[HTML]{EDEDED}
\textbf{+CRLD} & \textbf{72.08} & \textbf{74.59} & \textbf{78.22} & \textbf{76.49} & \textbf{75.71} & \textbf{75.48} & \textbf{75.43} \\
\cline{1-8}
 MLLD $\dagger$ \cite{mlld} & 72.19 & 74.11 & 77.08 & 76.63 & 75.35 & 75.18 & 75.09 \\
 \rowcolor[HTML]{EDEDED}
\textbf{+CRLD} $\dagger$ & \textbf{72.42} & \textbf{74.87} & \textbf{78.28} & \textbf{76.94} & \textbf{76.02} & \textbf{75.45} & \textbf{75.66} \\ \bottomrule \end{tabular} \label{tab:gen_study}
\end{table*}

\subsection{Main Results}
\noindent \textbf{Distillation performance}. 
We present extensive experimental results on CIFAR-100, Tiny-ImageNet, and ImageNet datasets using a diversity of teacher-student pairs in Tables ~\ref{tab:cifar100_homo} to \ref{tab:imagenet}. Specifically, the proposed CRLD outperforms all existing methods on all evaluated datasets across teacher-student pairs of both homogeneous (Tables \ref{tab:cifar100_homo}, \ref{tab:timagenet}, and \ref{tab:imagenet}) and heterogeneous (Tables \ref{tab:cifar100_het} and \ref{tab:imagenet}) architectures. When using MLLD's \cite{mlld} training configurations (marked with ``$\dagger$''), our method achieves further performance gains and leads MLLD by a considerable margin.  

\begin{figure}[t] \centering
\includegraphics[width=\linewidth]{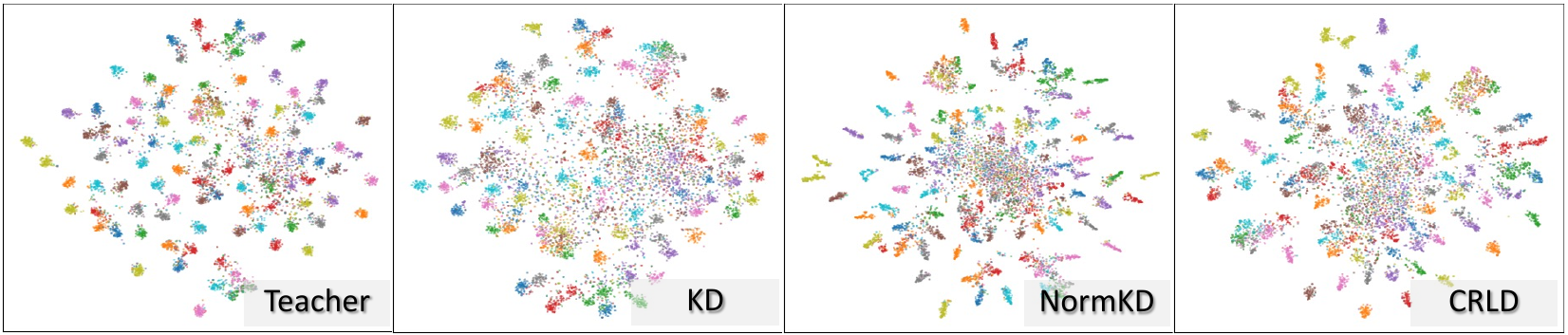}
  \caption{t-SNE visualisation of teacher's and distilled student's features on CIFAR-100.} \label{fig:tsne}  \vspace{-6pt} 
  \Description{t-SNE visualisation on the CIFAR-100 dataset.}
\end{figure}

\begin{figure}[t] \centering
\includegraphics[width=0.98\linewidth]{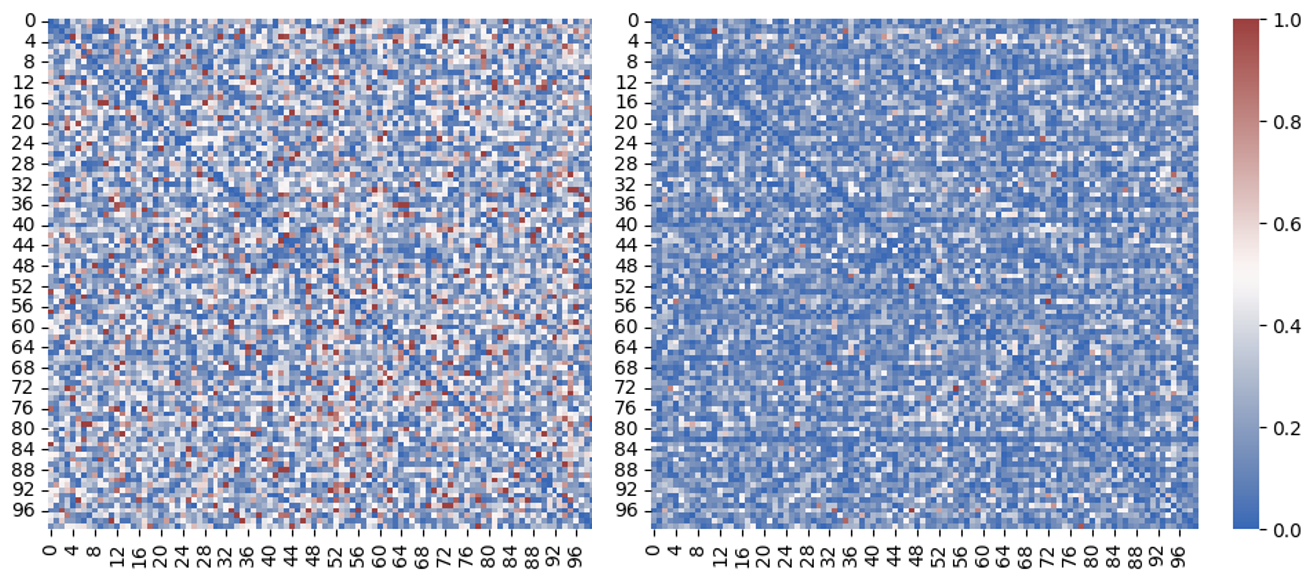} 
  \caption{Class-wise similarity maps between teacher and student predictions by NormKD and CRLD on CIFAR-100.} 
  \Description{Similarity maps between teacher and student predictions.} \label{fig:correlation_map} \vspace{-6pt}
\end{figure}

\noindent \textbf{Generalisation capabilities}
In addition to NormKD \cite{normkd}, we also apply the proposed CRLD to state-of-the-art logit-based knowledge distillation frameworks \cite{kd, nkd, mlld, dkd} and report the results in Table~\ref{tab:gen_study}. Note that for a fair comparison, we report our reproduced results for compared methods, using official implementations and specifications. The experimental results cogently validate the generalisation capability of our method. The proposed CRLD works orthogonally with existing knowledge distillation methods and can be easily incorporated to significantly boost knowledge transfer performance without introducing any extra network parameter or any additional inference overhead.

\subsection{Ablation Studies} \label{sec:ablation}
\noindent \textbf{Design of consistency regularisation.}
We break down our full training objective and investigate the play of each individual term in CRLD's overall effectiveness. A set of ablation experiments are conducted with results presented in Table~\ref{tab:abl}. First, we observe that within-view losses are individually effective and consistency within the strong view alone is more effective compared to weak view alone (Expt. A-B). Intriguingly, cross-view consistencies are harmful when used individually (Expt. B-C), but are rather beneficial when applied in concert with within-view consistencies (Expt. E-J). Finally, our ablation experiments (Expt. G-J) demonstrate that each individual consistency objective in our full objective plays a non-negligible part and their joint play leads to the optimal performance. Note that our experiments also highlight that \textit{the effectiveness of CRLD does not stem from a mere increase in the diversity of training samples}, as a notable $+1.56\%$ accuracy gain is achieved compared to when the exact same set of strong view augmentation policies are applied in a naive manner (\textit{i.e.}, Expt. B $\rightarrow$ J).

\begin{figure}[t] \centering
\includegraphics[width=\linewidth]{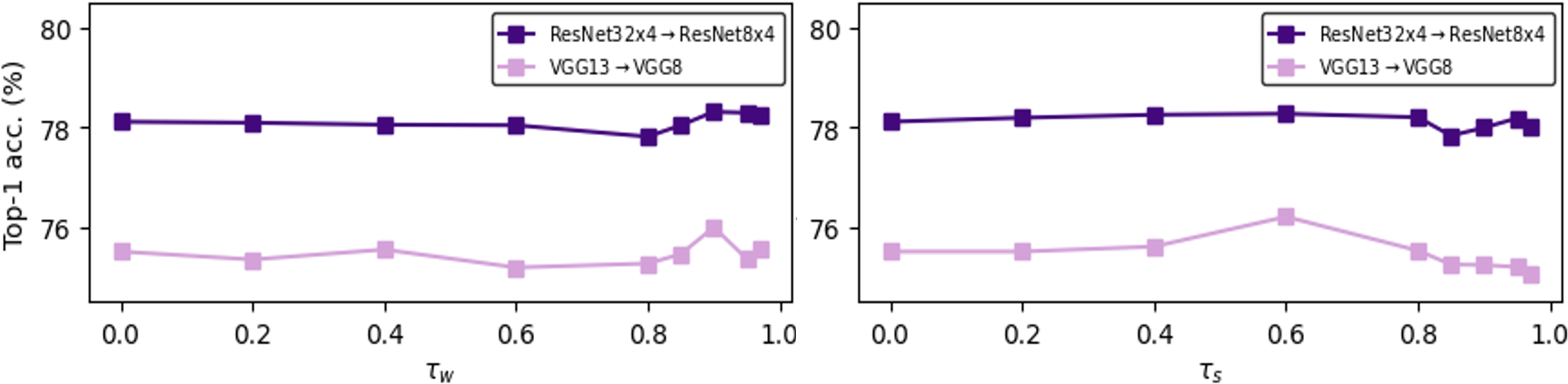} 
  \caption{Sensitivity of CRLD against varying $\tau_w$ and $\tau_s$ values on CIFAR-100.} \label{fig:tau_curve} \vspace{-8pt}
\end{figure}
\begin{figure}[t] \centering
\includegraphics[width=1.01\linewidth]{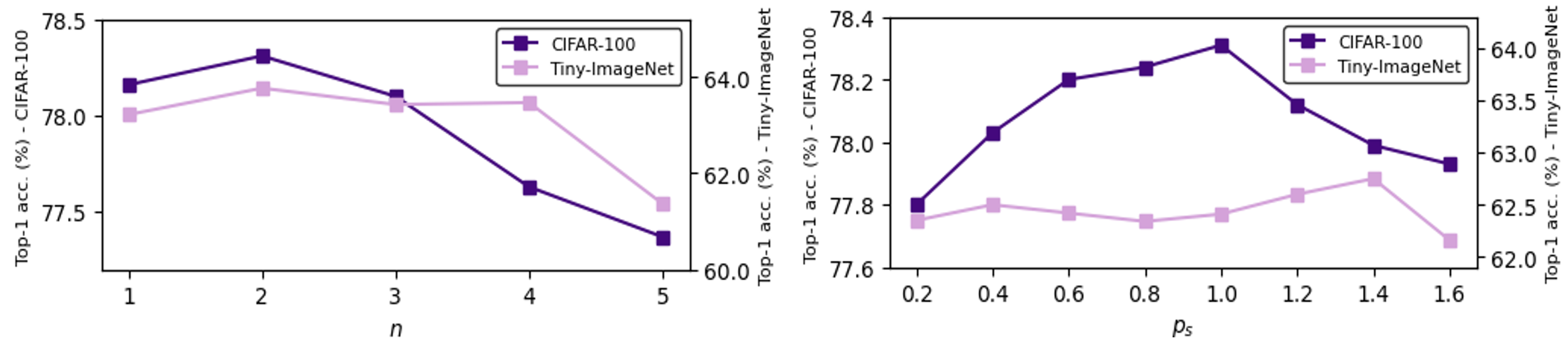}
  \caption{Sensitivity of CRLD against varying strengths of strong view transformation on CIFAR-100 and Tiny-ImageNet.} \label{fig:strength_curve} \vspace{-10pt}
\end{figure}

\noindent \textbf{Strengths of view transformations.} Table~\ref{tab:abl_trans_pair} probes how the absolute and relative strengths of CRLD's view transformations impact its performance. ``w/o CVL'' denotes ``without cross-view learning''. Apparently, CVL is beneficial regardless of the transformation strengths, but the ``strong-weak'' duo produces the best results. See Supplementary Materials for more discussions.

\noindent 
\textbf{Sensitivity to $\mathbf{\tau_w}$ and $\mathbf{\tau_s}$.} 
Figure \ref{fig:tau_curve} plots the performance of CRLD across different $\tau_w$ \& $\tau_s$ values. According to the empirical results, CRLD demonstrates limited sensitivity to these thresholding hyperparameters over the entire hyperaprameter space, despite peak performance within specific intervals.

\noindent 
\textbf{Sensitivity to strength of strong view transformations.}
To see how the strength of strong view transformations may influence the performance of CRLD, we conveniently tweak hyperparameter $n$, the number of view transformation operations randomly sampled and applied from all RandAugment operations. Moreover, we also apply a probability multiplier, $p_s$, to modulate the parameter values of sampled operations. Varying $n$ and $p_s$ allows an intuitive understanding of the impact of view transformation strength, plotted in Figure \ref{fig:strength_curve}, and the selection of these hyperparameters. More details can be found in the Supplementary Materials.

\begin{table}[t]
\caption{Effect of different view transformation strengths.}
 \center \small \captionsetup{skip=6pt} \tabcolsep=0.15cm
 \renewcommand{\arraystretch}{0.90}
  \begin{tabular}{c|cc} \toprule
 Method & CIFAR-100 & Tiny-ImageNet \\ \Xhline{3\arrayrulewidth} 
 w\slash o CVL & 76.26 &  60.83 \slash 83.08 \\
 Weak-Weak & 76.66 & 62.83 \slash 84.10 \\
 Strong-Strong & 76.73 & 61.67 \slash 83.94 \\
 \rowcolor[HTML]{EDEDED}
 \textbf{Strong-Weak} & \textbf{78.22} & \textbf{63.77} \slash \textbf{84.57} \\ \bottomrule 
\end{tabular} \label{tab:abl_trans_pair} \vspace{-1pt}
\end{table}

\subsection{Further Analyses}
\noindent \textbf{Training dynamics}. For further insights into the training profile of different methods, in Table~\ref{fig:acc_curve} we plot the evolution of training and test accuracies at each epoch throughout the training process. We observe that NormKD demonstrates much higher training accuracy than other methods but has only comparable or even lower test accuracy with respect to MLLD and DKD, which implies overfitting on training data. When the proposed CRLD is applied to NormKD, training accuracy lowers while test accuracy notably increases, suggesting alleviated overfitting and improved generalisation brought about by CRLD. In addition, we also notice less oscillatory test accuracy curves of CRLD, which is likely due to improved generalisation and mitigated confirmation bias of our method. 

\noindent \textbf{t-SNE visualisation}. We visualise the feature space learnt by the student using different logit-based distillation methods. As seen in Figure~\ref{fig:tsne}, features learnt using the proposed CRLD are significantly more seperable in the feature space, with more tightly clustered class-wise features and greater inter-class feature variations. These observations imply greater generalisation of the learnt model and substantiate the superiority of the proposed distillation method.

\noindent \textbf{Teacher-student output correlations.} To understand how well a trained student is able to mimic its teacher's predictions from a different perspective, we compute and visualise the correlations between student's and teacher's predictions in the Euclidean space in Figure~\ref{fig:correlation_map}. The left map corresponds to NormKD \cite{normkd} and the right CRLD applied to NormKD. It is clear that with CRLD, the average distance between teacher and student predictions are significantly reduced for all categories on the test data --- a compelling evidence of better distilled teacher knowledge and greater generalisation capabilities of the trained student.

\noindent \textbf{Distillation without ground-truths.} 
We assess the performance of different methods under the ``label-free knowledge distillation'' set-up, a more practical scenario where GT labels used to train the teacher are no longer available when performing KD. As shown in Table~\ref{tab:lfkd}, GT labels are indispensable to feature-based methods, and the proposed CRLD is the most resilient to missing GT labels amongst logit-based methods. 

\noindent \textbf{Application to ViT.}
To verify the effectiveness of our method on transformer-based models, we further consider the scenario where we distill from a ViT-L~\cite{vit} teacher to a ResNet-18 student. Table~\ref{tab:vit} presents the performance of different methods compared to CRLD on Tiny-ImageNet, where a 100-epoch training policy is employed. As observed, CRLD substantially outperforms all other methods, which suggests that our method generalises well to models with significantly distinctive underlying architectures.

\noindent \textbf{Implication on Beyer et al.~\cite{funmatch}.} Finally, we revisit the ``seemingly contradictory'' conclusions made in ~\cite{funmatch} which we have raised earlier on.
It turned out our conclusions do not contradict: The consistent matching in \cite{funmatch} is exactly our within-view consistency regularsation. \cite{funmatch} argues consistent matching alone outperforms inconsistent matching alone, which aligns with our observations (Table~\ref{tab:abl} A \& B \textit{v.s.} C \& D). Our work takes a step further by suggesting that consistent and inconsistent (\textit{i.e.}, cross-view) matchings are compatible, and can lead to state-of-the-art results when used in tandem.

\begin{table}[t] \centering \small \tabcolsep=0.10cm
\caption{Top-1 accuracy (\%) under the label-free knowledge distillation (LFKD) set-up on CIFAR-100.}
\renewcommand{\arraystretch}{0.90}
\begin{tabular}{cc|cc} \toprule
\multirow{2}{*}{\begin{tabular}[c]{@{}c@{}} Method \\ \end{tabular}} & Teacher & ResNet32$\times$4 & VGG13 \\
& Student & ResNet8$\times$4 & VGG8 \\
\Xhline{3\arrayrulewidth} 
\multirow{2}{*}{Feature KD}
& FitNets \cite{fitnets} &  1.39 & 1.09 \\
& OFD \cite{ofd} & 1.43 & 1.71  \\
\hline 
\multirow{4}{*}{Logit KD}
& KD \cite{kd} & 73.76 & 73.49 \\
& MLLD \cite{mlld} & 74.10 & 73.03 \\
& NormKD \cite{normkd} & 76.49 & 74.39 \\ 
& \cellcolor{sgray} \textbf{CRLD} & \cellcolor{sgray} \textbf{77.82} & \cellcolor{sgray} \textbf{75.36} \\
\bottomrule \label{tab:lfkd} \vspace{-10pt}
\end{tabular}
\end{table}

\begin{table}[t] \centering \tabcolsep=0.08cm \small
\caption{Top-1 accuracy (\%) on Tiny-ImageNet for ViT-L-to-ResNet18 distillation.}
\renewcommand{\arraystretch}{0.90}
\begin{tabular}{cc|ccccccc} \toprule
Teacher & Student & DKD & NKD & KD & NormKD & LSKD & MLLD & \cellcolor{sgray} \textbf{CRLD} \\ \Xhline{3\arrayrulewidth}
86.43 & 56.90 & 59.41 & 60.41 & 60.50 & 61.83 & 62.07 & 62.44 & \cellcolor{sgray} \textbf{63.41}  \\ 
\bottomrule 
\end{tabular} \label{tab:vit}
\end{table}

\noindent \textbf{More discussions.} More analyses and discussions are provided in the Supplementary Materials.

\section{Conclusion }
In this paper, we presented a novel logit-based knowledge distillation framework named CRLD. The motivation of CRLD lies in revamping popular ideas found in the semi-supervised learning literature, such as consistency regularisation and pseudo-labelling, to combat the overconfident teacher and confirmation bias problems in knowledge distillation. Our design of within-view and cross-view consistency regularsations, enabled by weak and strong image transformations and coupled with a confidence-based soft label selection scheme, leads to a highly effective and versatile knowledge distillation framework. Extensive experiments demonstrate that CRLD can boost existing logit-based methods by considerable margins and sets new records on different image classification datasets and under different configurations.

\newpage
\begin{acks}
This work was supported in part by NSFC (62322113, 62376156), Shanghai Municipal Science and Technology Major Project (2021SH\\ZDZX0102), and the Fundamental Research Funds for the Central Universities.
\end{acks}

\bibliographystyle{ACM-Reference-Format}
\bibliography{sample-base}

\end{document}


\title[Cross-View Consistency Regularisation for Knowledge Distillation - Supplementary Materials]{Cross-View Consistency Regularisation for \\ Knowledge Distillation - Supplementary Materials}

\author{Weijia Zhang}
\affiliation{
  \institution{Shanghai Jiao Tong University}
  \city{Shanghai}
  \country{China}}
\email{weijia.zhang@sjtu.edu.cn}

\author{Dongnan Liu}
\affiliation{
  \institution{University of Sydney}
  \city{Sydney}
  \country{Australia}}
\email{dongnan.liu@sydney.edu.au}

\author{Weidong Cai}
\affiliation{
  \institution{University of Sydney}
  \city{Sydney}
  \country{Australia}}
\email{tom.cai@sydney.edu.au}

\author{Chao Ma}
\authornote{Corresponding author.}
\affiliation{
  \institution{Shanghai Jiao Tong University}
  \city{Shanghai}
  \country{China}}
\email{chaoma@sjtu.edu.cn}

\maketitle
 
\vspace{50cm}
\section{List of Strong View Transformation Operations} \label{sec:strong_aug}
A full list of image transformation operations used for strong view augmentation in CRLD is given in Table~\ref{tab:randaug}. All transformations except for Cutout \cite{cutout} are part of the RandAugment strategy initially proposed in \cite{randaugment}. In CRLD, $n=2$ operations are randomly sampled from all 14 RandAugment transformation strategies, followed by Cutout.
The strength (\textit{i.e.}, the operation parameter) $v$ is set independently for each sampled operation and stochastically using the following equation:
\begin{equation} \label{eq:v}
    v = v_{\text{min}} + (v_{\text{max}} - v_{\text{min}}) * p
\end{equation}
where $v_{\text{min}}$ and $v_{\text{max}}$ are the the lower and upper bounds of the parameter range for corresponding operations in Table~\ref{tab:randaug}; $p \in [0,1]$ is a random number for stochastic parameter adjustment.
For Cutout, its parameter $v_{\text{co}}$ is generated by:
\begin{equation} \label{eq:vco}
    v_{\text{co}} = 0.5 \times p_{\text{co}}
\end{equation}
where $p_{\text{co}} \in [0,1]$ is another random number such that $v_{\text{co}} \in [0,0.5]$ always holds.

\begin{figure}[t] \centering
  \includegraphics[width=\linewidth]{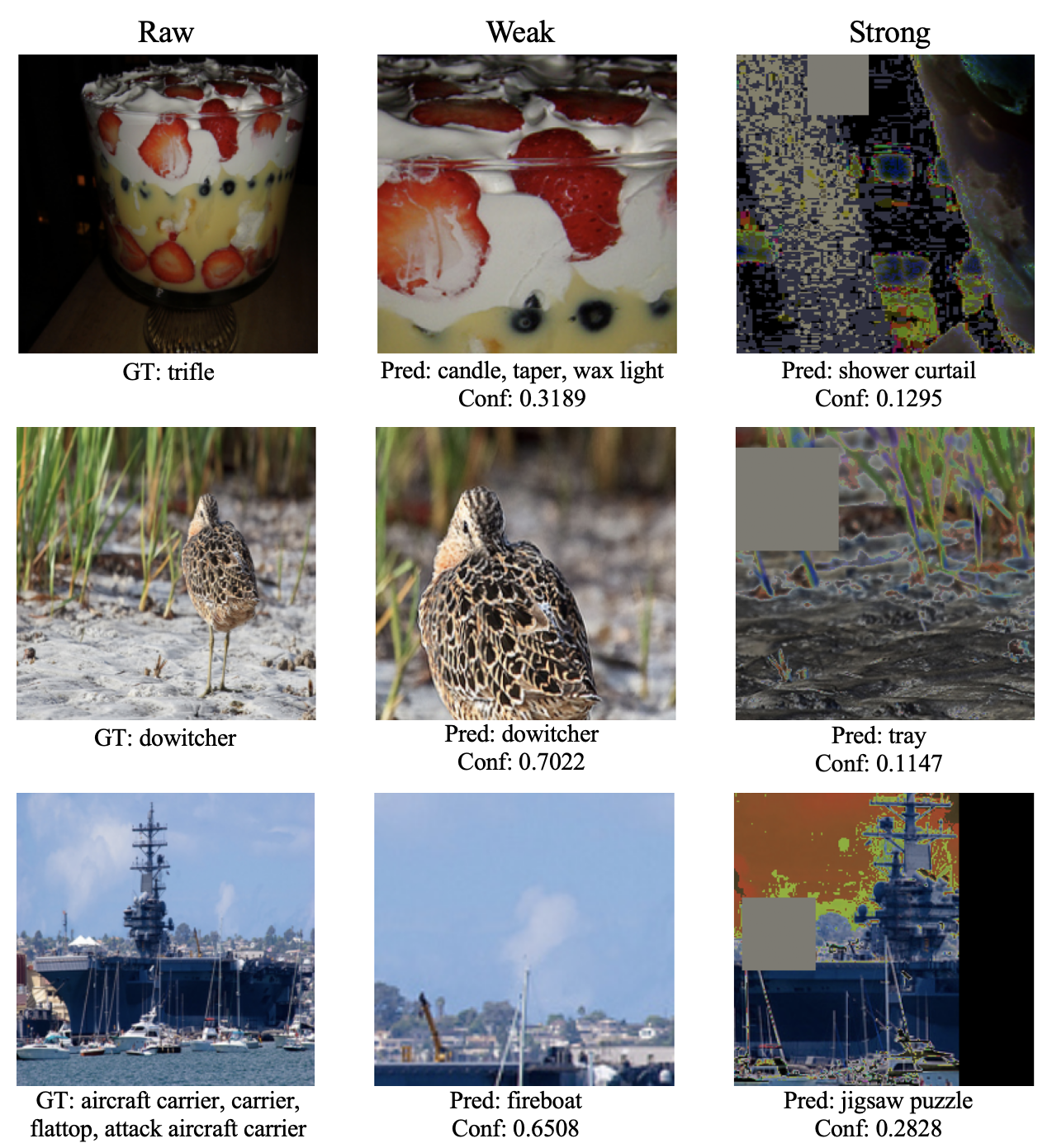}
  \caption{Examples of ImageNet \cite{imagenet} images transformed by the proposed weak and strong transformations and predictions made by a ResNet32$\times$4 teacher.}
  \Description{Similarity maps between teacher and student predictions.} \label{fig:aug_example}
\end{figure}

\begin{table*}[t] \centering
\caption{List of transformation operations used for strong view transformation.}
\begin{tabular}{cp{8cm}c} \toprule 
Transformation & \multicolumn{1}{c}{Description} & Param. Range \\ \Xhline{3\arrayrulewidth} 
\texttt{Autocontrast} & Automatically adjusts image contrast by setting the darkest pixel to black and lightest to white & - \\
\texttt{Brightness} & Adjusts image brightness & [0.05, 0.95]\\
\texttt{Color} & Adjusts image colour balance & [0.05, 0.95] \\
\texttt{Contrast} & Adjusts image contrast & [0.05, 0.095] \\
\texttt{Equalize} & Equalises image histogram & [0, 1] \\
\texttt{Identity} & Keeps image unchanged & [0, 1] \\
\texttt{Posterize} & Reduces number of bits for each image channel & [4, 8] \\
\texttt{Rotate} & Rotates image & [-30, 30] \\
\texttt{Sharpness} & Adjusts image sharpness & [0.05, 0.95] \\
\texttt{Shear\_x} & Shears image along horizontal axis & [-0.3, 0.3] \\
\texttt{Shear\_y} & Shears image along vertical axis & [-0.3, 0.3] \\
\texttt{Solarize} & Inverts all image pixels above a given threshold & [0, 256] \\
\texttt{Translate\_x} & Translates image horizontally & [-0.3, 0.3] \\
\texttt{Translate\_y} & Translates image vertically & [-0.3, 0.3] \\
\texttt{Cutout} & Sets pixels inside a random square patch within image to gray & [0, 0.5] \\
\bottomrule
\end{tabular}
\label{tab:randaug}
\end{table*}

\section{Further Analyses}
\noindent \textbf{Examples of challenging strongly-augmented images.}
In Figure~\ref{fig:aug_example}, we showcase some challenging examples produced by the proposed view transformation to support our motivation for a confidence-based soft label selection mechanism. As can be seen, both weak and strong views can be misclassified by a well-trained ResNet32$\times$4 teacher model. In particular, the strongly-augmented view can be extremely challenging and sometimes almost completely indiscernible. Cross-view consistency criteria forcibly imposed on these misleading predictions only serves to harm the student's learning, which is however alleviated by our proposed confidence-based soft label selection.   

\noindent \textbf{Strengths of view transformations.} The success of CRLD hinges on a pair of strongly- and weakly-transformed images to establish cross-view consistency regularisation objectives. It is of interest to investigate to what extent the absolute and relative strengths within a pair of view transformations impact the subsequent cross-view learning. To this end, we consider two additional set-ups: 1) using two independently composed weakly-transformed views (denoted as ``Weak-Weak''); 2) using two independently composed strongly-transformed views (``Strong-Strong''), and compare them against the original strong-weak consistency regularisation design (``Strong-Weak'') as well as the baseline set-up with no cross-view learning executed (``w\slash o CVL.''). 

From the results in Table 9, we easily draw the following conclusions: 1) Any form of cross-view learning, despite different view transformation strengths, leads to performance gains over the baseline, which again substantiates the effectiveness of our proposed cross-view consistency regularisation. 2) A pair of view transformations of identical strength results in degraded performance compared to the proposed ``strong-weak'' learning. We attribute this to additional dark information mined and transferred across two different spaces of transformed images, compared to limited knowledge dug within a single space. 3) When using transformation pairs of the same strength, it is not decisive what strength level may be more beneficial --- this may be dataset- and task-dependent.

\noindent \textbf{Sensitivity to $\mathbf{\tau_w}$ and $\mathbf{\tau_s}$.} The confidence-base soft label mining mechanism essentially features a quantity-quality trade-off. With a higher threshold, we demand soft labels of higher quality but an inevitably smaller number of them are selected for knowledge transfer; with a lower threshold, we have richer knowledge in the form of teacher's soft labels involved in the knowledge transfer, but their quality and reliability are lower on average. 
Figure 6 visualises such a trade-off by plotting the performance of CRLD against different values of $\tau_w$ and $\tau_s$. 

We notice that the optimal trade-off point for $\tau_w$ is at a higher value. This is expected since the predicted confidence for the less challenging weakly-augmented view is much higher on average, which means a sufficient number of soft predictions of the teacher fall within the top-confidence interval. As such, setting a high $\tau_w$ ensures soft-labels are selected in high quality while also in ample quantity. By contrast, most teacher predictions for the strong view are less confident. A much smaller $\tau_s$ is required to ensure sufficient teaching signals. 

\noindent \textbf{Sensitivity to strength of strong view transformations.} 
Following previous investigations, we further carry out a set of experiments to probe into the impact of strong view transformation in different strengths on CRLD's performance. First, we vary $n$, the number of view transformation operations randomly sampled and applied sequentially from all RandAugment operations in Table~\ref{tab:randaug}. As shown in the left sub-plot in Figure 7, a combination of more strong view transformation operations degrades the performance of CRLD. This is expected since with an increasingly challenging strongly-augmented view, the teacher struggles to provide correct and beneficial soft predictions, and the student could be misled by a predominant amount of distracting and harmful signals from the teacher. Although the value of $n$ can be tweaked for each dataset and even for each teacher-student configuration for further performance gains, we simply use $n=2$ by default for simplicity. 

To enable fine-grained control over the strength of strong augmentations (\textit{i.e.}, RandAugment operations and Cutout), we also introduce a probability multiplier $p_s$ to tune the parameter value of each operation (listed in Table~\ref{tab:randaug}). $p_s$ is introduced into Equations~\ref{eq:v} and ~\ref{eq:vco} as:
\begin{equation} \label{eq:v}
    v = v_{\text{min}} + (v_{\text{max}} - v_{\text{min}}) * p * p_s
\end{equation}
and
\begin{equation} \label{eq:vco}
    v_{\text{co}} = 0.5 \times p_{\text{co}} * p_s
\end{equation}
Note that a higher $p_s$ value does not necessarily mean stronger transformation for all operations listed in Table~\ref{tab:randaug}. Nevertheless, larger $p_s$ leads to more strongly-transformed images on average, and we are interested in finding out how sensitive our method is to changes in the parameter values. From the right sub-plot in Figure 7, we notice that the performance indeed varies with changing $p_s$. Overly large or small $p_s$ tends to produce inferior performance, which echoes our findings in Table 9 and the above experiments on different $n$ values. Besides, different datasets are observed to manifest different sensitivity patterns to $p_s$. More fine-grained control of the transformation parameters are left for future work. 

\begin{table*}[t] \centering \small
\begin{minipage}{0.60\textwidth} \centering  \tabcolsep=0.03cm
\captionof{table}{Top-1 accuracy (\%) under the label-free knowledge distillation (LFKD) set-up on CIFAR-100.} 
\begin{tabular}{cc|cccccc|c} \toprule
\multirow{4}{*}{\begin{tabular}[c]{@{}c@{}} Method \\ \end{tabular}} & \multirow{2}{*}{Teacher}  & ResNet56 & ResNet110 & ResNet32$\times$4 & WRN-40-2 & WRN-40-2 & VGG13 &
\multirow{4}{*}{\begin{tabular}[c]{@{}c@{}} Avg. \\ \end{tabular}} \\
&  & 72.34 & 74.31 & 79.42 & 75.61 & 75.61 & 74.64 \\
& \multirow{2}{*}{Student} & ResNet20 & ResNet32 & ResNet8$\times$4 & WRN-16-2 & WRN-40-1 & VGG8 \\
& \space  & 69.06 & 71.14 & 72.50 & 73.26 & 71.98 & 70.36 & \\
\Xhline{3\arrayrulewidth} 
\multirow{2}{*}{Feature KD}
& FitNets \cite{fitnets} & 1.04 & 1.01 & 1.39 & 1.14 & 1.09 & 1.09 & 1.13 \\
& OFD \cite{ofd} & 2.09 & 1.13 & 1.43 & 1.49 & 2.27 & 1.71 & 1.69 \\
\hline 
\multirow{4}{*}{Logit KD}  
& KD \cite{kd} & 70.66 & 73.53 & 73.76 & 74.79 & 73.41 & 73.49 & 73.27 \\
& MLLD \cite{mlld} & 70.88 & 72.54 & 74.10 & 74.88 & 72.56 & 73.03 & 73.00 \\
& NormKD \cite{normkd} & \textbf{71.36} & \textbf{74.35} & 76.49 & 76.04 & 74.82 & 74.39 & 74.58 \\
& \cellcolor{sgray} \textbf{CRLD} & \cellcolor{sgray} 71.18 & \cellcolor{sgray} 74.20 & \cellcolor{sgray} \textbf{77.82} & \cellcolor{sgray} \textbf{76.53} & \cellcolor{sgray} \textbf{75.10} & \cellcolor{sgray} \textbf{75.36} & \cellcolor{sgray} \textbf{75.03} \\
\bottomrule \label{tab:lfkd}
\end{tabular} \label{tab:crfd}
\end{minipage}
\hfill
\begin{minipage}{0.39\textwidth} \centering \tabcolsep=0.05cm \vspace{-10pt}
\captionof{table}{Towards feature-space consistency regularisation.} 
\begin{tabular}{c|cc} \toprule
 & ResNet32$\times$4 & VGG13 \\ & ResNet8$\times$4 & VGG8 \\ \Xhline{3\arrayrulewidth} 
 KD (\texttt{logits}) & 73.53 & 73.43 \\
 \rowcolor[HTML]{EDEDED} \textbf{+CRLD (\texttt{logits})} & \textbf{77.60} & \textbf{75.27} \\ \hline
 FitNets (\texttt{pool-feat}) & 76.74 & 73.87 \\
 \rowcolor[HTML]{EDEDED} \textbf{+CRLD (\texttt{pool-feat})} & \textbf{77.73} & \textbf{75.43} \\ \hline
FitNets (\texttt{feats-3}) & 73.66 & \textbf{72.27} \\
 \rowcolor[HTML]{EDEDED} \textbf{+CRLD (\texttt{feats-3})} & \textbf{73.92} & 70.78 \\ \hline
  FitNets (\texttt{feats-2}) & \textbf{73.45} & \textbf{72.30} \\
  \rowcolor[HTML]{EDEDED} \textbf{+CRLD (\texttt{feats-2})} & 73.17 & 70.37 \\ \bottomrule
\end{tabular} \label{tab:crfd}
\end{minipage}
\end{table*}

\noindent \textbf{Label-free knowledge distillation.}
Given that consistency regularisation has been a widely successful technique in semi-supervised learning, one may naturally ponder whether the proposed CRLD would work with solely unlabelled training samples. 
Herein, we define a slightly deviating KD task coined ``label-free knowledge distillation (LFKD)'', which forbids the use of ground-truth labels during knowledge transfer. LFKD is highly relevant and practical scenario --- it is often the case that we have access to only a pre-existing, pre-trained teacher model, but the annotations used to train the teacher are no longer accessible (annotations can be costly or not made public due to privacy or commercial concerns, especially in industrial contexts).
In Table~\ref{tab:lfkd}, we re-evaluate several methods under the LFKD setting. We find that feature-based methods fail completely under LFKD, which is due to a vague causal linkage between feature mimicking and the downstream classification task. In contrast, logit-based methods deliver comparable performance. CRLD remains the top performer, which signifies its resilience to the absence of annotations. Besides, their performance may be even stronger when using a more knowledgeable teacher with high-accuracy predictions that can largely supersede the role of ground-truth labels. Note that logit-based methods such as DKD and NKD do not support LFKD due to the involvement of ground-truth labels in their objective formulation.

\noindent \textbf{Feature-space consistency regularisation.}
Thus far, our CRLD has been strictly logit-based, with the consistency objectives enforced in the logit space.
By extension, we are curious about to what extent can our proposed regularisation schemes be extended into the feature space. 
In theory, closer to the network's input end, features' level of abstraction lowers, and the discrepancy between their representations grows. Forcing feature maps to match can therefore hurt student training within its intermediate stages. Therefore, intuitively we expect degraded student performance as consistency regularisation moves towards shallower layers.

Next, we outline our experiment set-ups. Most feature-based methods adopt a convolutional regressor layer to adapt the student feature to the teacher feature. For a fair performance comparison, we follow this practice by employing two such layers, one for student's predictions of the weakly-augmented view and the other for the strongly-augmented view. 
Notation-wise, we use $\texttt{pool-feat}$ to denote the pooled feature map by average pooling, right before the Softmax layer; we use $\texttt{feats-i}$ to denote the feature map produced by the $i\text{th}$ feature blocks immediately after the activation layer. Please refer to the \textit{mdistiller} codebase for details. As for the loss function, following FitNets, we adopt the mean squared error (MSE) loss in place of the original Kullback-Leibler divergence (KLD) loss for our consistency regularisation objectives. 

The experimental results (Table~\ref{tab:crfd}), corroborate our earlier intuition. As consistency regularisation moves upper stream --- from the logit layer to the pooled feature layer (\texttt{pool-feat}), and then to the 3rd and 2nd feature layers (\texttt{feats-3} \& \texttt{feats-2}), an overall decrease in performance is observed (detailed descriptions found in Supplementary Materials). Interestingly, the optimal performance is reached when consistency regularisation is applied to \texttt{pool-feat} instead of the logit space. Nevertheless, the largest performance gains are yielded when it is applied in the logit space, suggesting that the capacity of consistency regularisation is maximally utilised when working with logits. Motivated by these results, we leave the question of how to unleash the potential of the proposed method within the feature space for future work.

\noindent \textbf{Difference to CS-KD~\cite{cs-kd}.} CS-KD is a method that involves self-knowledge-distillation and consistency regularisation. CRLD differs from CS-KD in at least three fundamental aspects: 1) Motivation: While both consider overconfidence, CS-KD tackled intra-class variation whereas we used teacher’s privileged knowledge to reduce confirmation bias, a key pain point CS-KD fails to address. 2) Nature of regularisation: CS-KD uses self-KD whereas we
use cross-agent matching; 3) Choice of matched pairs: CS-KD uses class-wise samples whereas we use transformed views instance-wise. A fundamental flaw of CS-KD is that the non-target dark knowledge, proven crucial in KD, may entirely mismatch between a pair in their formulation (only target knowledge guaranteed to match).

\bibliographystyle{ACM-Reference-Format}
\bibliography{sample-base}